\documentclass{article}

\usepackage[nonatbib, final]{neurips_2019}

\usepackage[utf8]{inputenc} 
\usepackage[T1]{fontenc}    
\usepackage{hyperref}       
\usepackage{url}            
\usepackage{booktabs}       
\usepackage{amsfonts}       
\usepackage{nicefrac}       
\usepackage{microtype}      
\usepackage{graphicx}
\usepackage{subcaption}
\usepackage{color}
\usepackage{amsmath}

\title{AlignNet: Unsupervised Entity Alignment}

\author{
  Antonia Creswell, Kyriacos Nikiforou, Oriol Vinyals \\
  \textbf{Andre Saraiva, Rishabh Kabra, Loic Matthey, Chris Burgess,}\\ \textbf{Malcolm Reynolds, Richard Tanburn,  Marta Garnelo, Murray Shanahan} \\
  DeepMind
}

\begin{document}

\maketitle

\begin{abstract}

Recently developed deep learning models are able to learn to segment scenes into component objects without supervision. This opens many new and exciting avenues of research, allowing agents to take objects (or entities) as inputs, rather that pixels. Unfortunately, while these models provide excellent segmentation of a single frame, they do not keep track of how objects segmented at one time-step correspond (or align) to those at a later time-step. The alignment (or correspondence) problem has impeded progress towards using object representations in downstream tasks. In this paper we take steps towards solving the alignment problem, presenting the AlignNet, an unsupervised alignment module.

\end{abstract}

\section{Introduction}

Many every day tasks require us to interact with objects over extended periods of time, during which objects enter and leave our field of view. This may be due to our head or body movements, because we interact with them or because the objects are moving due to other causal effects. Despite this, we are still able to keep track of object identity across time, even through long term occlusion.

\begin{figure}[h!]
\centering
    \includegraphics[width=\textwidth]{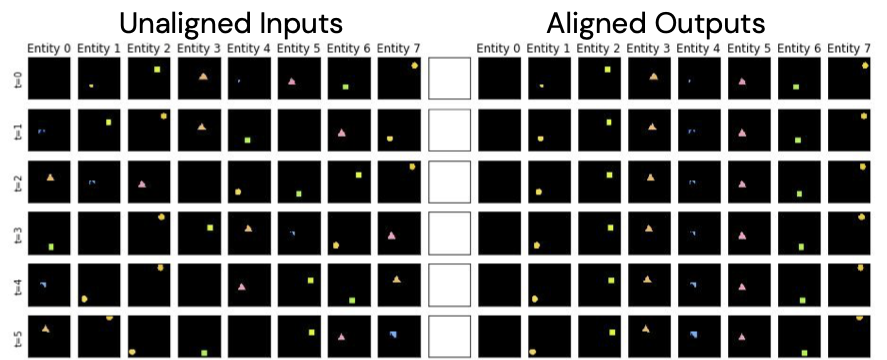}
    \caption{On the left `Unaligned inputs', entities switch column between time-steps. On the right `Aligned outputs', each column contains the same object across time in its new position (these results were obtained using AlignNet).}
    \label{fig:main_results}
\end{figure}

While our interactions with the world often require us to focus on objects, when training agents we commonly use pixel inputs (e.g. images). Since we, as humans, break our observations down in to objects (or entities) \cite{spelke2007core}, one natural avenue for exploration would be to use entity (or object) level inputs in our models \cite{janner2018reasoning, watters2019cobra, veerapaneni2019entity, reyes2019learning}.

MONet \cite{burgess2019monet} and other \cite{greff2019multi} unsupervised segmentation models, provide an exciting opportunity to train agents (and their transition models e.g. COBRA \cite{watters2019cobra}) on lower dimensional object representations rather than images. Unfortunately, current object-based transition models \cite{janner2018reasoning, watters2019cobra, veerapaneni2019entity, reyes2019learning} lack a crucial ability to understand how objects segmented in one frame correspond with those segmented in another frame. This makes it more difficult to integrate information about objects over time or even compute losses at the entity level, because correspondence between predicted and target objects are unknown. In this paper we propose the AlignNet, a model capable of computing correspondence between objects across time, not just from one time-step to the next, but across long sequences.

For the majority of the paper, we concentrate on fully observable environments, aligning currently observed entities with those observed in the previous time-step. However, we will also show results in a partially observable environment, aligning current inputs with an object-based memory instead. By incorporating an object-based memory, we create an inductive bias for object persistence; once a new object appears it must continue to exist even if it disappears for some time. This allows the model to not only deal with appearance of new entities and disappearance of entities, but also the reappearance of entities through longer term occlusion. 

The AlignNet has two key components; first is a dynamics model that predicts updates in the representation of the entities aligned in the previous time-step, to match the representations of the entities received in the current
one. The second is a permutation model that permutes entities at the current time-step to correspond with the order of the previously aligned entities. The dynamics model helps to bridge the difference between the representation of the entities at the current and previous time-step.

\section{Why is this problem important?}

\begin{figure}
\centering
    \includegraphics[width=\textwidth]{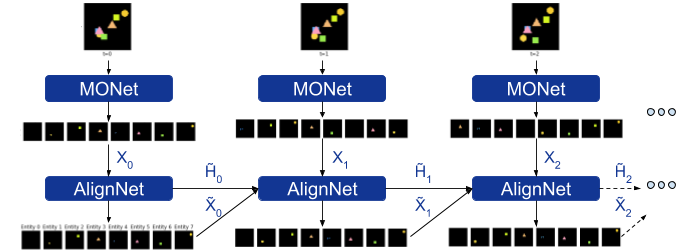}
    \caption{AlignNet is used to align MONet \cite{burgess2019monet} entities across multiple time-steps.}
    \label{fig:monet_align_net}
\end{figure}
\begin{figure}
\centering
    \includegraphics[width=\textwidth]{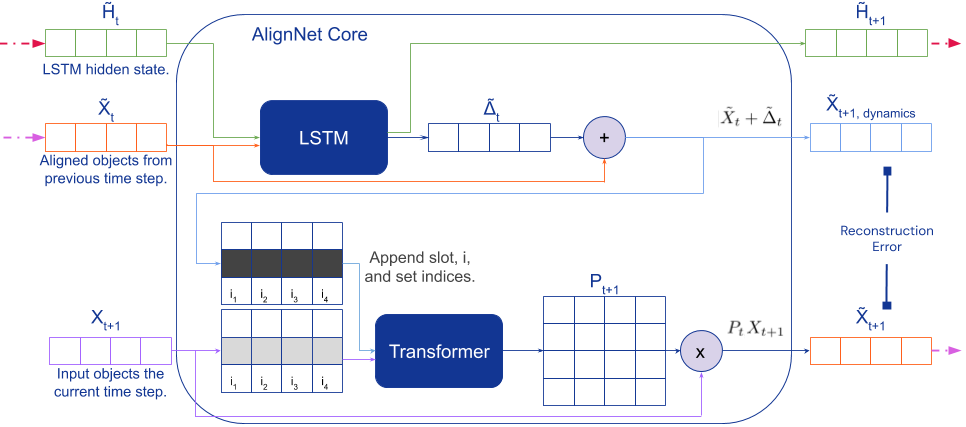}
    \caption{\textbf{The AlignNet Core}. The LSTM acts as the dynamics model predicting where the previously aligned entities, $\tilde{X}_t$, would be at the next time-step, $\tilde{X}_{t+1, dynamics}$. The transformer predicts a permutation matrix, $P_{t+1}$ to align the current input, $X_{t+1}$ with the previously aligned entities, $\tilde{X}_t$. Applying the permutation matrix to the input entities gives the aligned entities at the current time-step, $\tilde{X}_{t+1}$.}
    \label{fig:align_net_core}
\end{figure}

In this section we demonstrate the need for alignment in existing entity-based models, not only for learning dynamics \cite{chang2016compositional, yi2019clevrer, riochet2020occlusion}, but also for object based planning \cite{janner2018reasoning, watters2019cobra, veerapaneni2019entity, reyes2019learning} and in robotics \cite{ferreira2019learning}.

To train models we need to be able to compute losses between predictions and targets. Problems arise when both the output of our model (the prediction) and the target is a set of objects, because the correspondence between the predicted and the target objects is unknown and so we are not able to compute losses.

There are two ways to solve this problem. Firstly, if the inputs and targets are fed to the model in some consistent (or canonical) order then the model can easily exploit this ordering and learn to make predictions in this same consistent (or canonical) order. Note that a canonical order suggests some higher level rules by which the objects are ordered (e.g. based on objects' locations in a scene) and a consistent order suggests that the $n^{th}$ input entity or predicted entity always corresponds to the same "token" or instance of that entity over time. Secondly, if the inputs and targets are not in a canonical (or consistent) order, the model cannot learn to output predictions in a canonical or consistent order. Instead we need to know correspondences between the prediction and the targets. The first requires the alignment of the inputs and targets over time and the second requires alignment between the predictions and the targets.

Many approaches are currently avoiding the alignment problem by either taking losses in pixel space (Section \ref{sec:avoid}) or using privileged information to align inputs and targets (Section \ref{sec:privileged}). Few papers, when faced with the entity alignment problem, use weak heuristics to implicitly align a set of predicted objects with a set of target objects in order to compute losses (Section \ref{sec:hack}), but this is often not treated in the main body of the paper. Creswell et al. \cite{creswell2020alignnet} and Piloto et al. \cite{piloto2020learning} are among the first to acknowledge the alignment problem although Piloto et al. also use privileged information to align inputs and targets in their entity-based dynamics models. However, a similar problem exists in the computer vision literature and is referred to as re-identification \cite{ye2020deep, inbook}. Unlike our model that learns without supervision, models trained to perform re-identification often require access to ground truth labels and bounding boxes \cite{zhan2020simple}.

\subsection{Avoiding the alignment problem by taking losses in pixel space}\label{sec:avoid}

Both Janner et al. \cite{janner2018reasoning} and Watters et al. \cite{watters2019cobra} obtain losses for their entity-based transition models by mapping from their predicted entities' features to full image scenes in pixel space to take losses. Similarly, Riochet et al. \cite{riochet2020occlusion} learn to map from a set of objects to a segmentation and depth map and take losses in pixel space. 

The problem with mapping back to image space is that it is not always possible to do so, due to lack of a decoder model, and losses may be more semantically meaningful if taken in the entity representation space. Additionally, it can be computationally expensive to apply large decoder models such as MONet to decode back to image space.

\subsection{Using privileged information to provide inputs and targets in a consistent order across time.}\label{sec:privileged}

Chang et al. \cite{chang2016compositional} avoid the alignment problem by using a ground truth state space where the inputs and targets are in a consistent order and the associations between the input entities and target entities are known. 

Janner et el. \cite{janner2018reasoning} look at planning in object space, computing an object wise $l_2$ distance, using privileged information. They \cite{janner2018reasoning} encode objects in a consistent order by using ground truth consistently ordered masks in their perception model. Similarly, Ferreira et al. \cite{ferreira2019learning} also use ground truth masks to provide their objects in a consistent order across time and to ensure that the targets are in the same order.

Yi et al. \cite{yi2019clevrer} make use of `ground-truth motion traces and event histories of each object in the videos' allowing them to use an $l_2$ loss to train their models, avoiding the alignment problem. 

\subsection{Attempts at computing entity-wise losses.}\label{sec:hack}

Veerapaneni et al.'s \cite{veerapaneni2019entity, reyes2019learning} `entity cost' involves computing a minimum pair-wise loss between entity masks. Their approach requires entities to be mapped back to pixel (mask) space before computing their losses. Mapping back to pixels provides a strong positional bias when aligning objects since the loss between entities will only be small if they overlap significantly in pixel space. Similarly, Smith et al. \cite{smith2019modeling}, use a handcrafted method to compute losses based on intersection over union of object masks.

The problem with these approaches is threefold. Firstly, they do not take advantage of lower dimensional representations that may encode objects in interesting (e.g. disentangled) ways and secondly, they require a decoder, which may not always be available. Finally, when two objects become very close (or occlude one another), position is not enough to resolve which object is which, rather we need additional information about the dynamics of the objects to resolve the alignment.

One algorithm often exploited to compute correspondence between two sets of entities is the Hungarian. Given an adjacency matrix, the Hungarian algorithm solves the minimum assignment problem between two sets of entities. However, the Hungarian assumes access to an adjacency matrix populated with the costs of matching up each of the entities in one set with those in the other. The adjacency matrix may be approximated by the mean-squared-error between entity representations but this approach is not sufficient to deal with partially observable environments or ambiguous cases where dynamics is needed to resolve alignment as we will show in Section \ref{sec:exp}. Finally, the Hungarian algorithm is non-differentiable and therefore we cannot pass gradients through it for training.

To conclude this section, alignment may be useful not only for computing losses but also for integrating information about objects in observations across time. Consider a scene with two objects moving with constant velocity. If you know how objects correspond between frames it is easy to compute the velocity, if not, this becomes much harder. Alignment computes this correspondence across time.

Solving alignment would relieve the need for privileged information and lift many of the limitations faced by these models. It is also worth noting that most of these problems are still concerned with fully observable environments and so in this paper, we too focus on fully observable environments to serve the community with a solution as soon as possible. We present some initial results in partially observable environments.

\section{The AlignNet Model}

Given a sets of object observations, $\mathcal{X}_{\tau=t} = \{x_1, x_2, x_3, ..., x_N \}_{\tau=t}$ across time, $\tau$, we would like to predict how each object at the current time-step, $x_i \in \mathcal{X}_{\tau=t+1}$, corresponds with each object in the previous time-steps, $x_j \in \mathcal{X}_{\tau \leq t}$. Here $x$ is the visual representation of an object that may change when an object moves due to lighting and other conditions.

In this paper we consider aligning objects across many time-steps. We can achieve this by first looking at how to align objects across a single time-step and then recursively applying alignment across all time-steps. To being with, we concatenate the elements in each set of objects at times $\tau=t$ and $\tau=t+1$ to obtain lists, $X_t$ and $X_{t+1}$ which have a fixed order, unlike sets:

\[X_{t} = [x_1, x_2, x_3, ..., x_N ]_{t}\]
\[X_{t+1} = [x_{j_1}, x_{j_2}, x_{j_3}, ..., x_{j_N} ]_{t+1}\]

\subsection{Problem Setup} \label{sec:set_up}

We would like to learn a function, $f$, (see Equation \ref{eqn:init}) that finds an alignment between $X_{t+1}$ and $\tilde{X}_t$, where $\tilde{X}$ is $X$ re-arranged in some consistent order (as shown on the right-hand side of Figure \ref{fig:main_results}). Initially, in this paper, we assume a full observable environment and so objects visible at $t=0$ are the same objects visible at $t>0$, therefore we choose to define the order of objects to be the order at $t=0$, resulting in $\tilde{X}_0=X_0$. To find an alignment it may be necessary to use information provided by previously aligned inputs, $\tilde{H}_t = [ \tilde{X}_0, \tilde{X}_1, ..., \tilde{X}_{t-1}]$ to help resolve ambiguities in alignment.

The distribution of aligned entities at the next time-step, $\tilde{X}_{t+1}$, given the entities at the current time-step, $X_t$, and the history, $\tilde{H}_t$, can be formulated as the conditional distribution in Equation \ref{eqn:normal}, where we assume the distribution to be a Gaussian with mean given by Equation \ref{eqn:init} and variance, $\sigma^2$.

\begin{equation} \label{eqn:normal}
    p(\tilde{X}_{t+1} | X_t, \tilde{H}_t) = \mathcal{N}(f(X_t, \tilde{H}_t), \sigma^2)
\end{equation}

\begin{equation} \label{eqn:init}
\tilde{X}_{t+1} = f(X_t, \tilde{H}_t)
\end{equation}

The choice of function, $f$, is critical and should capture two things: (1) the permutation (or re-ordering) of elements in $X_t$ and (2) $f$ must take into account how object appearance may change over time; $f$ must account for the dynamics of objects across time.  Therefore we choose $f$ to consist of a permutation, $P$, and a linear approximation for the dynamics of the objects, allowing a vector $\tilde{\Delta}$ to capture the changes between time-steps. The function $f$ is defined in Equation \ref{eqn:simple_f} where $\tilde{\Delta}$ depends on $\tilde{H}_t$. Note that $\tilde{X}_t = P_t X_t$.

\begin{equation} \label{eqn:simple_f}
    f(X_t, \tilde{H}_t) = P_t X_t + \tilde{\Delta}_t
\end{equation}

Plugging function, $f$, (Equation \ref{eqn:simple_f}) in to Equation \ref{eqn:normal} we obtain:

\begin{equation} \label{eqn:normal_final}
    p(\tilde{X}_{t+1} | X_t, \tilde{H}_t) = \mathcal{N}(PX_t + \tilde{\Delta}_t, \sigma^2)\pi(P)\pi(\tilde{\Delta}_t)
\end{equation}

where $\pi(P)$ and $\pi(\Delta_t)$ are the prior distributions over $P$ and $\Delta_t$ respectively. Since there is no preference for $P$, we may choose a uniform prior over all possible permutations, therefore $\pi(P)=\pi_P$ is a constant. Again, recall  that $\tilde{\Delta}_t$ depends on $\tilde{H}_t$.

The evidence lower bound (ELBO) for Equation \ref{eqn:normal_final} is then given by Equation \ref{eqn:elbo}:
\begin{equation} \label{eqn:elbo}
\begin{split}
    \log p(\tilde{X}_{t+1} | X_t, \tilde{H}_t) &\geq \mathbb{E}_{q(P, \tilde{\Delta}_t| X_t, \tilde{H}_t)} \log p(\tilde{X}_{t+1} | P, X_t, \tilde{\Delta}_t, \tilde{H}_t) \\ &- \mathbb{KL}[q(P|X_t, \tilde{\Delta}_t)||\pi_P] \\ &- \mathbb{KL}[q(\tilde{\Delta}_t|\tilde{H}_t)||\pi(\tilde{\Delta}_t)]
\end{split}
\end{equation}

We choose to factor the posterior as follows, $q(P,\tilde{\Delta}_t|X_t, \tilde{H}_t) = q(P|X_t, \tilde{\Delta}_t)q(\tilde{\Delta}_t |\tilde{H}_t)$. Factorising the posterior in this way, the deltas depend only on the history, $\tilde{H}_t$, and not on the current input. This limits the information that is available to the deltas and helps to avoid trivial solutions. One trivial solution, that we avoid, would be the permutation matrix being an identity matrix and the delta accounting for all of the difference.

The choice of prior, $\pi(\tilde{\Delta}_t)$ is more difficult than for $\pi(P)$. If we assume that object representations do not change very much over time we may assume a Gaussian prior, $\mathcal{N}(0, 1)$, however, this assumption may not always hold. Alternatively, we could assume a uniform distribution over all possible delta values, however, it is possible that we may obtain trivial solutions.

\subsection{Implementation}

The input to our model, $X_t$, is a list of representation vectors extracted from an image observation at time, $t$, using a pre-trained MONet \cite{burgess2019monet} model (see Figure \ref{fig:monet_align_net}). When extracting entities we use ground truth masks, which we shuffle to ensure that the input object representations, $X_t$, are in a random order, where we know the order only for evaluation purposes.

We approximate the remaining distributions (all distribution except the priors) in Equation \ref{eqn:elbo} as follows.

\begin{itemize}
    \item $q(\tilde{\Delta}_t|\tilde{H}_t)$ is approximated as a Gaussian distribution. An LSTM is used to predict the mean and standard deviation of the deltas at each time-step given the aligned input, $\tilde{X}_t$, from the current step along with the LSTM state, $\tilde{H}_t$.
    \item $q(P|X_t, \tilde{\Delta}_t)$ is the output of a transformer applied to both the output of the dynamics model, $\tilde{X}_t + \tilde{\Delta}_t$, and the (unaligned) inputs in the next step, $X_{t+1}$. Rather than using the final values predicted by the transformer, we use the similarity (or attention) matrix as the output. We then apply the Sinkhorn algorithm \cite{mena2018learning}, a differentiable algorithm that operates on a matrix to give a soft approximation of a permutation matrix.
    \item $p(\tilde{X}_{t+1} | P_t, X_t, \tilde{\Delta}_t)$ is parameterised as a Gaussian distribution, $\tilde{X}_{t+1} \sim \mathcal{N}(P_t X_t + \tilde{\Delta}_t, \sigma)$ with fixed variance, $\sigma=1$ for simplicity.
\end{itemize}

Throughout this paper, for additional simplicity, we use the means of the Gaussian distributions rather than samples since we found that using the means had no adverse effect on the results. When evaluating the expectation, $\mathbb{E}_{q(P, \tilde{\Delta}_t| X_t, \tilde{H}_t)} \log p(\tilde{X}_{t+1} | P_t, X_t, \tilde{\Delta}_t)$, it is important to note that we are always aligning the next step, $X_{t+1}$, with the current aligned step, $\tilde{X}_t$, and we assume that $X_0 = \tilde{X_0}$. This is equivalent to learning a $\tilde{\Delta}^*_t$ and $P^*_t$ such that $\tilde{X}_t = {P^*_t}^T X_{t+1} - \tilde{\Delta}^*_t$, under some constraints, such as, $P^*_t$ is a permutation matrix. The expectation is then given by the mean-squared difference between $\tilde{X}_t + \tilde{\Delta}_t$ and $P_{t}X_{t+1}$. The loss is given in Equation \ref{eqn:loss}, note that the first $\mathbb{KL}$ term simplifies to the entropy, $\mathbb{H}$, of the permutation matrix. Our model is illustrated in Figure \ref{fig:align_net_core}.

\begin{equation} \label{eqn:loss}
    loss = \|\tilde{X}_t + \tilde{\Delta}_t - P_{t}X_{t+1} \|_2^2 + \beta_1  \mathbb{H}(P_{t}) + \beta_2  \mathbb{KL}[q(\tilde{\Delta}_t|\tilde{H}_t)||\pi(\tilde{\Delta}_t)]
\end{equation}

\section{Experiments and Results}\label{sec:exp}

\begin{figure}[h!]
\centering
    \includegraphics[width=\linewidth]{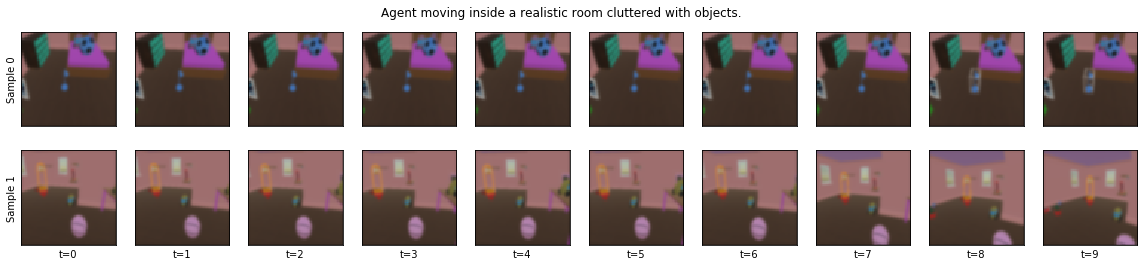}
    \caption{Observations of an agent interacting in an simulated 3D Unity Room environment \protect\cite{Hill2020Environmental, das2020probing, hill2020human} filled with objects from 58 different classes in 3 different sizes and 10 different colour.}
    \label{fig:room}
\end{figure}

We demonstrate the AlignNet's performance on 5 tasks (illustrated in Figures \ref{fig:sw}, \ref{fig:blai:perm} and \ref{fig:room}) spanning the three environments: SpriteWorld \cite{spriteworld19}, Physical Concepts \cite{piloto2018probing, piloto2020learning} and Unity Room, a 3D partially observable environment \cite{Hill2020Environmental, das2020probing, hill2020human}. Finally, we show some additional results in Unity Room (Section \ref{sec:improved_align_net}) and Physical Concepts (Section \ref{sec:improved_align_net_blai}) using a modified version of the AlignNet that incorporates memory to deal with partially observable environments.

\subsection{SpriteWorld}
 SpriteWorld \cite{spriteworld19} is a 2D environment made up of 2D shapes with continuous colours, sizes, positions and velocities. In this paper, we use three shapes: squares, triangles and circles. When two sprites `collide' in SpriteWorld one object occludes the other and the sprites continue to move with the same velocity they had before.
 
\begin{figure}[h!]
\centering
    \begin{subfigure}[c]{\textwidth}
        \includegraphics[width=\linewidth]{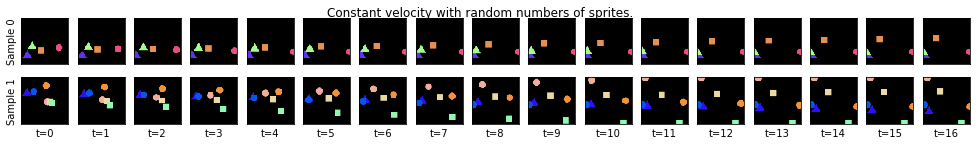}
        \caption{\textbf{Task (a):} The number of sprites is drawn from a uniform distribution, $U\{1, 7\}$, each sprite moves with constant random velocity and stops at the edges, becoming partially occluded. The colour and shape of each object is sampled uniformly.}
        \label{fig:sw:const_v}
    \end{subfigure}
    \begin{subfigure}[c]{\textwidth}
        \includegraphics[width=\linewidth]{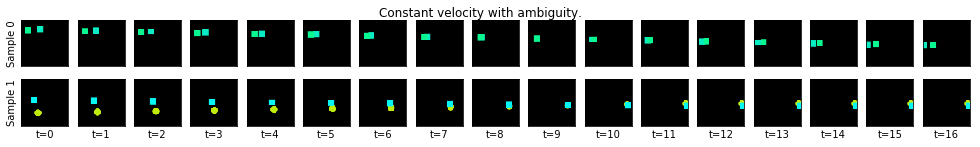}
        \caption{\textbf{Task (b):} Each example contains two sprites, in $50\%$ of examples sprites have the same shape and colour and in $90\%$ of examples sprites collide with constant velocity, between time-step $t=5$ and $t=10$. Sprites stop at the edges.}
        \label{fig:sw:const_v_amb}
    \end{subfigure}
    \begin{subfigure}[c]{\textwidth}
        \includegraphics[width=\linewidth]{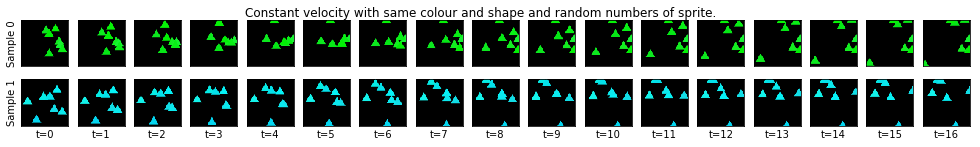}
        \caption{\textbf{Task (c):} The number of sprites is drawn from a uniform distribution, $U\{1, 7\}$. The sprites move with constant random velocity and stop at the edges. All sprites in each sample are the same shape and colour. Sprites stop at the edges.}
        \label{fig:sw:const_v_same}
    \end{subfigure}
\caption{Datasets generated in the SpriteWorld environment.}
\label{fig:sw}
\end{figure}

 We begin by testing AlignNet on three tasks in the SpriteWorld, the tasks are described in Figure \ref{fig:sw}. In task (a) we test how well the AlignNet can handle up to seven objects moving with random constant velocity. While task (a) tests if AlignNet can handle many objects, it is possible that the model learns to match objects based only on their visual properties and not based on their velocities. In tasks (b) and (c) we create tasks with ambiguities, that can only be resolved if the model understands dynamics.
 
 In task (b), $45\%$ of the time, the model is presented with entities of the same shape and colour colliding. At the point where the entities collide, it would be impossible for a model that did not capture dynamics to resolve which object was which after the collision. Therefore task (b) tests whether the AlignNet is indeed learning the correct dynamics. Task (c) takes this one-step further having up to seven objects of the same shape and colour, moving with constant velocity.
 
 In Table \ref{tab:res:sw} we compare the AlignNet to results using the Hungarian. The Hungarian is a non-differentiable algorithm that is used to solve the minimum assignment problem in the case where an adjacency matrix is given. For the comparisons we present in this work, we compute the adjacency matrix using the mean-squared-error between all pairs of object representations. Additional visual results are shown in Figures \ref{fig:SW_task_a_results}, \ref{fig:SW_task_b_results} and \ref{fig:SW_task_c_results}. Results for the first five (of 16) time-steps for task (b) are also shown in Figure \ref{fig:main_results}. We see that the AlignNet solves all tasks well, with some errors in task (c). On inspection, we found that the model fails on task (c) in some cases where more than two objects collide and where those objects have similar trajectories. This fail case would also be hard for humans to resolve.

 \begin{table}[h!]
     \centering
     \begin{tabular}{l| ccc|c|c|}
         & \multicolumn{3}{|c|}{Sprite World Task} & Physical Concepts & Unity Room \\
        & (a) & (b) & (c) & Continuity & (agent following policy) \\
        \hline 
        AlignNet accuracy & 100$\%$ & 100$\%$& 99.8$\%$& 100$\%$&  86.2$\%$  \\
        Hungarian  & 99.7$\%$ & 96.4$\%$ & 97.4$\%$& 98.9$\%$& 85.8$\%$ \\
        \end{tabular} 
     \caption{AlignNet performance (three significant figures) on each Task.}
     \label{tab:res:sw}
 \end{table}

\subsection{Physical Concepts: Continuity} \label{sec:blai_pillars}

\begin{figure}[h!]
\centering
    \includegraphics[width=\linewidth]{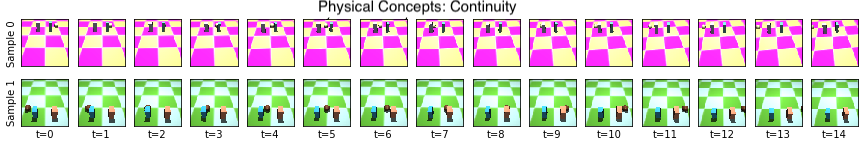}
    \caption{Samples from the Physical Concepts: Continuity dataset. In $75\%$ of examples the a ball rolls behind two pillars, in all other cases the ball rolls in front of the pillars.}
    \label{fig:blai:perm}
\end{figure}

We also demonstrate AlignNet's performance on an experiment inspired by Spelke et al. \cite{baillargeon1985object} that tests infants' understanding of object persistence. We use the "object persistence" task, demonstrated in Figure \ref{fig:blai:perm}, taken from the Physical Concepts task suite \cite{piloto2018probing} where a ball rolls behind two pillars $75\%$ of the time. While the ball is behind the pillar it cannot be seen, which tests AlignNet's ability to deal with short term occlusion. Additionally, the visual properties of the ball change as it moves, due to the effect of the different lighting conditions. The viewing angle is also shifted randomly during the observation while still focusing on the centre of the scene.

Our model achieves $100\%$ accuracy on this task, while the Hungarian achieves $98.9\%$ accuracy. Figure \ref{fig:blai_results} is a visualisation of AlignNet's performance on the Physical Concepts task, we see that the AlignNet is able to re-assign an object to the correct slot even after occlusion. If the AlignNet had not been able to deal with occlusion it would have placed the ball in a different slot after occlusion; rather the AlignNet knows that the ball is the same object before and after occlusion, assigning it to the same slot.

Although our models do very well on SpriteWorld and Physical Concepts tasks, the tasks are by no means easy. Many versions of the model did not work. For example, without entropy regularisation (the second term in the loss, Equation \ref{eqn:loss}) on the permutation matrix, $P_{t}$, the accuracy would peek and then start to go down as the model found ways to exploit softer assignments to minimise the objective during training. We found that $\beta_1=0.1$ worked well. Empirically, we also found that we did not need to use the $\mathbb{KL}[q(\tilde{\Delta}_t | \tilde{H}_t) || \pi(\tilde{\Delta}_t)]$ term in the loss (Equation \ref{eqn:loss}), reducing the number of hyper-parameters that need to be tuned.

The AlignNet also learns very low entropy solutions which means that hard and soft alignment are very similar. Using soft alignment allows gradients to be passed back through the model which may be useful when using the AlignNet for downstream tasks.

\subsection{Unity Room} \label{sec:unity_room}

In the previous tasks, the observations have be acquired by a stationary agent observing either a 2D scene from a static view (SpriteWorld) or a 3D scene from a camera, making very small movements left and right (Physical Concepts). In the Unity Room task the observations are collected by an agent following a learned policy where the agent was trained to pick up and move specific objects through a language instruction. This means that unlike the previous observations, these observations include examples of appearance, disappearance and re-appearance of objects as the agent switches its focus from one part of the room to another. The environment has the added complexity of containing many objects from 58 different classes in varying sizes and colours. 

We show visualisation of the AlignNet's performance on the Unity Room dataset in Figure \ref{fig:room_results}. The AlignNet achieves $86.2\%$  alignment accuracy, the Hungarian algorithm achieves similar performance. While our model is able to deal with the variety of object classes, colours, size and with some the agent's motion, our model is unable (by design) to deal with partially observable environments. When designing our model, we make an explicit assumption that the objects visible at $t=0$ are visible for the next steps (Section \ref{sec:set_up}) and therefore if an object disappears the model may not be able to handle this well. An example of this failure case is shown in Figure \ref{fig:room_results_fail}; the second slot, `Entity 2' is initially assigned a table, but once the table disappears from view at $t=7$ it is replaced by a similarly coloured object that was in a similar position before it disappeared.

\begin{table}[h!]
\centering
    \begin{tabular}{l|  c | c |}
        & Physical Concepts Free-Form & Unity Room (agent turning) \\
        \hline
        Memory AlignNet accuracy & 90$\%$ & 96$\%$ \\
        Hungarian accuracy & 62$\%$ & 81$\%$ \\
    \end{tabular} 
    \caption{Comparing performance of the Memory AlignNet (Section \ref{sec:improved_align_net} and \ref{sec:improved_align_net_blai}) to the Hungarian on the Physical Concepts Free-Form task (Section \ref{sec:improved_align_net_blai}) and the Unity Room task where the agent is turning (Section \ref{sec:improved_align_net}).}
     \label{tab:hard}
 \end{table}

What is significant though is that in some cases our model \textbf{can} deal with new objects appearing, this is demonstrated in `Entity 8' of Figure \ref{fig:room_results}, where a small pink shape appears from behind the white object that the agent is interacting with at $t=2$.

\subsection{Unity Room with the improved Memory AlignNet.} \label{sec:improved_align_net}

For these experiments we modified the AlignNet to have a slot-wise object-based memory and align with respect to the memory rather than with respect to the previous time-step (see Figure \ref{fig:memory_align_net}). We refer to this improved version of the AlignNet as \textit{Memory AlignNet}. We also make the dynamics model action conditional. By incorporating an object-based memory, we create an inductive bias for object persistence; once a new object appears it must continue to exist even if it disappears for some time. This allows the model to not only deal with appearance of new objects and disappearance of objects, but also the reappearance of objects. 

We create a modified task in the Unity Room to exhibit many examples of appearance, disappearance and reappearance of entities (or objects). In this modified task the model receives a sequence of 12 frames in which an agent situated in a Unity Room environment turns left for a number of steps drawn from the uniform distribution, $U\{6, 9\}$ and turns back for the rest of the time-steps. This ensures that the dataset captures objects moving in and out of view.

\begin{figure}
    \centering
    \includegraphics[width=\textwidth]{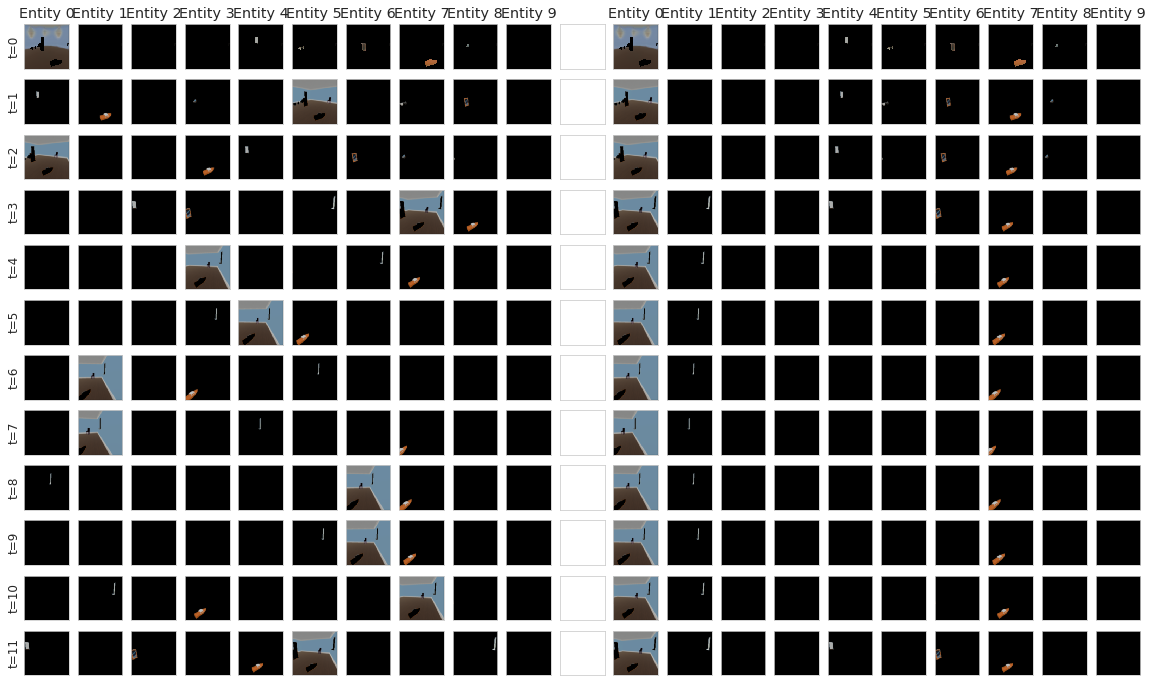}
    \caption{Visual Results on the Unity Room task where an agent is turning left for a number of steps and then turning right for the remainder of the steps. The unaligned inputs to the model are shown on the left and the aligned outputs are shown on the right. These results demonstrate that the Memory AlignNet is able to deal with new objects appearing (Entity 1 @ t=3) and is able to deal with entities disappearing (Entity 4 and 5 @ t=3) and reappearing in the correct slot (@ t=11). It also shows Entity 7 persisting in the same slot across time.}
    \label{fig:result:unity_room_spin}
\end{figure}

Our Memory AlignNet achieves $90\%$ accuracy on this task demonstrating the ability to deal with longer term occlusion over multiple frames as shown in Figure \ref{fig:result:unity_room_spin}. The Hungarian algorithm achieves only $62\%$ accuracy, this is lower than in the previous section because in this task there are more examples of appearance and re-appearance.

\subsection{Physical Concepts Free-Form with the Memory AlignNet.} \label{sec:improved_align_net_blai}

\begin{figure}[h!]
\centering
    \includegraphics[width=\linewidth]{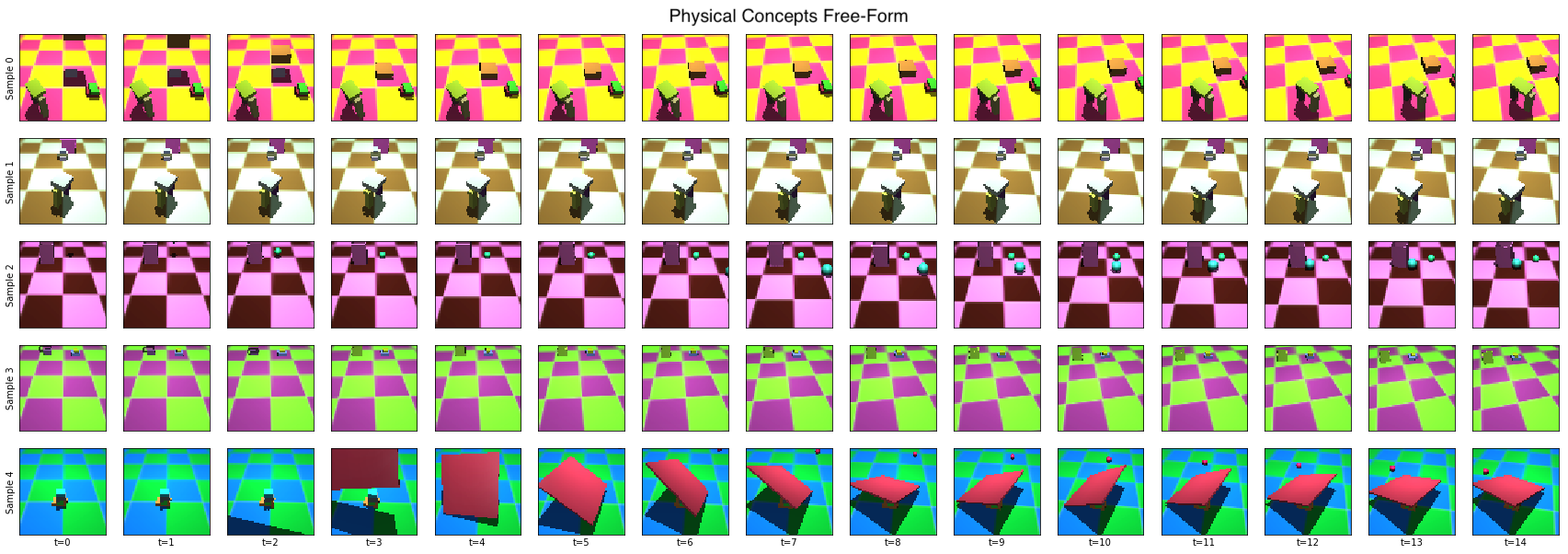}
    \caption{Samples from the Physical Concepts Free-Form environment  where objects roll, drop, collide and get occluded by other objects including containers. This dataset exhibits objects appearing, disappearing and reappearing later \protect \cite{piloto2020learning}.}
    \label{fig:blai_free_from}
\end{figure}

For these experiments we also use the Memory AlignNet (as in Section \ref{sec:improved_align_net}) and we use a more complex Physical Concepts \cite{piloto2020learning} task shown in Figure \ref{fig:blai_free_from}. In this task the model receives a sequence of 15 frames. Each frame contains a number of objects including balls, containers, planks and cuboids that interact with one another. Interactions include collisions, occlusions and containment events. Initially unseen objects may drop or roll into the agent's field of view. Both this and the variety of physical interactions make this a challenging task for the Memory AlignNet.

\begin{figure}
    \centering
    \includegraphics[width=0.7\textwidth]{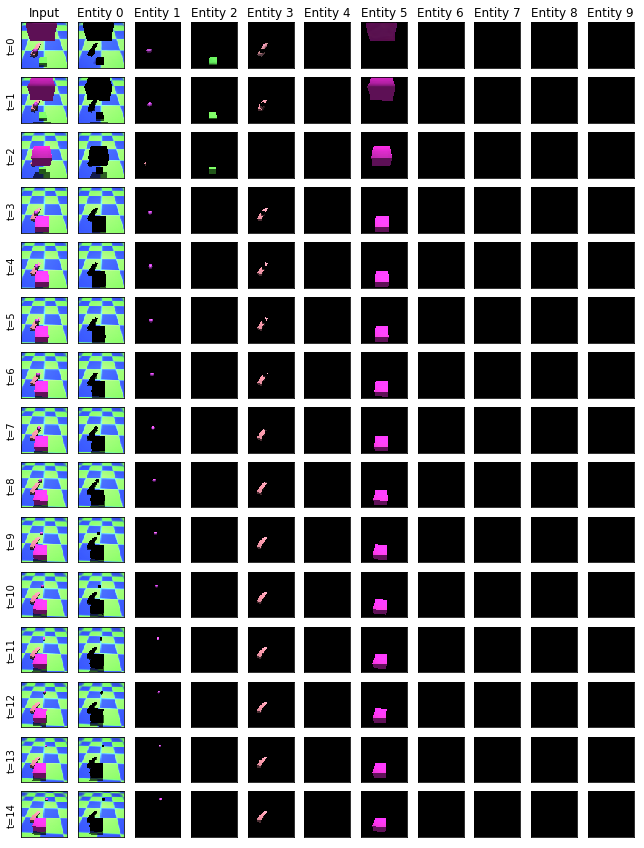}
    \caption{Visual results on the Physical Concepts Free-Form task. The first column shows the input at each time-step. The rest of the columns shows entities 0 to 9. We see that each entity is stable across time. At $t=2$ we see a small plank (Entity 3) being occluded by the purple box (Entity 5) and then re-appearing in the correct slot at $t=3$. We also see that the Memory AlignNet is able to keep track of the purple ball (Entity 1) rolling down the plank, even when it is partially occluded by the purple box (Entity 5) at $t=2$.}
    \label{fig:result:blai_free_form_results}
\end{figure}

While the Hungarian baseline achieves $81\%$ accuracy, our Memory AlignNet achieves $96\%$ accuracy. The Memory AlignNet does better than the Hungarian because it uses a dynamics model to deal with changes in lighting and viewing conditions and has a memory that helps the model to deal with re-appearance. The Hungarian algorithm has neither of these; it does not take in to account the dynamics of the objects and does not use memory. Visual results are shown in Figure \ref{fig:result:blai_free_form_results}.

\subsection{Summary of Results}

The AlignNet is a differentiable model that has learned to align entities without supervision while performing at least as well as the Hungarian algorithm, a hand-crafted and non-differentiable baseline, in fully observable environments and significantly outperforming when modified to deal with partially observable environments. We expect our model to be better than the Hungarian algorithm in several cases where there is  ambiguity that can only be resolved by understanding dynamics. In fact the SpriteWorld tasks that the Hungarian performs worst on are SpriteWorld Task (b) and (c), which have the most ambiguities. Further, we see that the Hungarian algorithm fails in partially observable environments, while our Memory AlignNet performs well.

\section{Related Work}

\textbf{Advantages of our approach over existing approaches.} 

Unlike Veerapaneni et al. \cite{veerapaneni2019entity, reyes2019learning} and Smith et al. \cite{smith2019modeling} our model performs an explicit alignment in the entity space that does not require a decoder model to map back to entities in pixel space or to the masks. Additionally, our model learns to use dynamics to resolve ambiguous cases where two objects may be visually similar and where there is occlusion.

While \cite{yi2019clevrer} ensure that a `combination of the three attributes uniquely identifies one object', we specifically look at datasets where scenes contain many examples of objects that are the same colour and shape (see Figure \ref{fig:sw}).

We have also significantly built on our earlier work, the Self-supervised Alignment Module \cite{creswell2020alignnet}, by incorporating a dynamics model. This allows our new version of the AlignNet to deal well with moving objects, changes in lighting conditions and view point.

\textbf{How do humans identify and individuate objects?}

Psychologists and cognitive scientists have tried to explain how humans keep track of objects (or entities) as they go in and out of their visual field. Pylyshyn \cite{Pylyshyn1989TheRO} proposes one explanation which he refers to as ``sticky indices''; when a new object first appears it is assigned an index which ``sticks'' to the object as it moves; similar to tracing an object with your finger. Pylyshyn does not give an exact mechanism by which these indices ``stick'' to objects (or entities). Some works suggest that we evaluate the gap between two representations and determine whether that gap is plausible; whether it can be explained by our model of the world. If the gap can be explained then we consider these entities (or objects) to have the same ``index''. This is very similar to how the AlignNet works, using the dynamics model to predict where objects should be and using the permutation model to perform the matching.

\textbf{Traditional and Deep Learning approaches to solving combinatorial problems.}

Alignment is different to the minimum assignment problem encountered in combinatorics because the minimum assignment problem assumes access to a similarity or adjacency matrix. This is true of both the Hungarian algorithm -- a traditional non-differentiable solution to the minimum assignment problem -- and deep learning approaches \cite{bello2016neural, vinyals2015pointer, milan2017data}, which both operate on an adjacency (or similarity) matrix. The AlignNet does not require a pre-defined similarity matrix. Rather, the dynamics model in the AlignNet learns to account for possible differences in the object representations allowing us to compute errors to train the AlignNet. Additionally, we consider a more general assignment problem where there may be no match, for example, if an object appears for the first time.

Andrychowicz \& Kurach \cite{andrychowicz2016learning} propose a model for sorting and merging sequences. However their proposed method is non-differentiable, unlike the AlignNet which is. ShuffleNet proposed by Lyu et al. \cite{lyu2019autoshufflenet}, shows that deep networks are able to predict permutation matrices. However, neither of these works focus on alignment. 

\textbf{Deep Learning Approaches to Object Tracking.}

It is important to note that our work is significantly different to traditional object tracking, in that we focus on keeping track of pre-extracted entities without assumed access to the extraction process, while most object tracking literature focuses on object tracking in images where the objects are not yet extracted. Additionally, we do not assume access to ground truth bounding boxes (or labels) and train our model without supervision.

An important and novel feature of the Memory AlignNet (Section \ref{sec:improved_align_net}) is its ability to deal with appearing, disappearing and re-appearing entities (or objects). He et al. \cite{he2018tracking} propose a model for tracking objects, but unlike the improved version of the AlignNet, their model cannot deal with reappearing objects because it terminates trackers when an object disappears. Additionally, He et al.  \cite{he2018tracking} assume that all objects in the sequence were visible at $t=0$, meaning that the model cannot deal with new objects appearing later in the sequence. This is an assumption that we were able to relax with the improved AlignNet. Further, while the improved AlignNet is able to account for new objects current works treat objects that are not sufficiently similar to those seen before as false detections \cite{yoon2019data}.

As in the Memory AlignNet, Valmadre et al. \cite{valmadre2017end} and  Yang \& Chan \cite{yang2018learning} also incorporate memory into their object tracking model. However, they only track a single object, while we use the AlignNet to keep track of multiple objects.

\textbf{Object-based reasoning in deep learning.}

Some progress has been made towards object (or entity) based models both in reinforcement learning \cite{zambaldi2018relational, kulkarni2019unsupervised} and in relational reasoning over objects \cite{yi2018neural, janner2018reasoning, ferreira2019learning, reyes2019learning}. These models show promise over models trained on raw-pixel inputs, but in general these works focus on fully observable environments where objects do not come in and out of view.

Several models exist that learn to extract objects (or entities) without supervision \cite{burgess2019monet, greff2019multi, nash2017multi, greff2016tagger}. However, if these models are applied to successive frames of a video, the output is a set of objects at each time-step and the correspondence between objects across time is unknown, especially in partially observable environments. This is the purpose of the AlignNet, with the Memory AlignNet able to keep track of entities in partially observable environments.

\textbf{Writing objects to memory.}

\begin{figure}
    \centering
    \includegraphics[width=\linewidth]{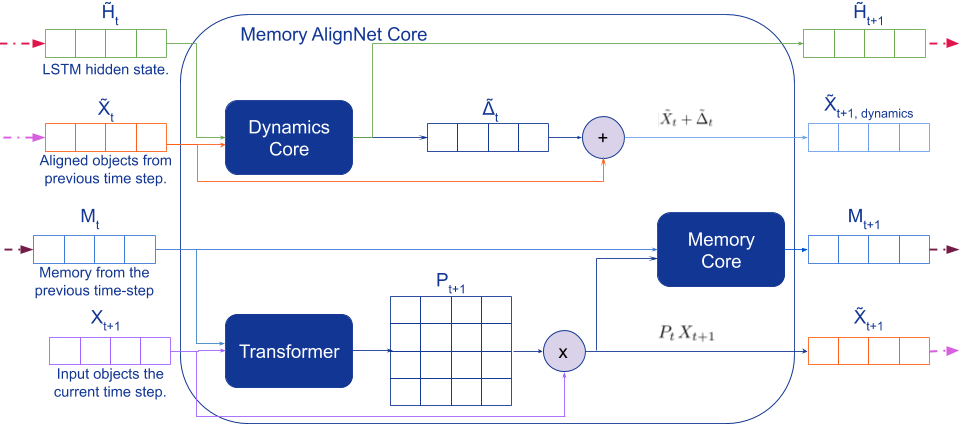}
    \caption{\textbf{The Memory AlignNet Core}. The improved version of the AlignNet with memory. The Memory Core is a slot-wise LSTM; where an LSTM is applied independently to each slot. The Memory Core takes the aligned entities $\tilde{X}_{t+1}$ as input as well as the memory, $M_t$ and the action taken (if any).}
    \label{fig:memory_align_net}
\end{figure}

Learning to write discrete entities into a slot-based memory can be hard because networks often use soft addressing mechanisms \cite{graves2014neural}. One exception to this is the Neural Map \cite{parisotto2017neural}, where observations are stored in a 2D spatial map based on the agent's location in an environment, they store a single representation of their observation rather than entities. In our improved AlignNet we incorporate a slot-based memory that allow us to keep track of discrete entities over time. We achieve this by applying a slot-wise LSTM to the aligned elements at each time-step, treating each slot independently and allowing each slot to accumulate the history of a single object (when the model is trained correctly). See Figure \ref{fig:memory_align_net} for details.

\section{Conclusions}

The AlignNet performs very well in fully observable environments, both 2D SpriteWorld and 3D Physical Concepts: Continuity. The model is able to learn to leverage dynamics to resolve ambiguous cases. For example, when two objects with similar shapes and colour collide, the AlignNet uses the entities' (or objects') dynamics to resolve which is which. On the Physical Concepts: Continuity task we demonstrated that the model can deal with short term occlusion, realistic lighting conditions and small changes in viewing angle.

For tasks in partially observable environments, we augmented the AlignNet with a slot-based memory (Section \ref{sec:improved_align_net}), which we refer to as Memory AlignNet. We found that Memory AlignNet significantly outperformed baselines in both the Unity Room environment and on the Physical Concepts Free-Form data, dealing well with the appearance of new entities and disappearance and re-appearance of entities. There is still work to be done to improve Memory AlignNet, namely by working on the architectures of the dynamics and memory models.

By providing a solution to the alignment problem, the AlignNet opens up many new and interesting opportunities for future work using objects in reinforcement learning and other downstream tasks.

\subsubsection*{Acknowledgments}
We would like to acknowledge Peter Battaglia, David Barrett and Danilo Rezende for their helpful discussions and feedback. We would also like to thank Phoebe Thacker and Lucy Campbell-Gillingham for their support.

\bibliography{neurips_2020_template}
\bibliographystyle{abbrv}

\appendix

\begin{figure}[h!]
    \centering
    \includegraphics[width=\linewidth]{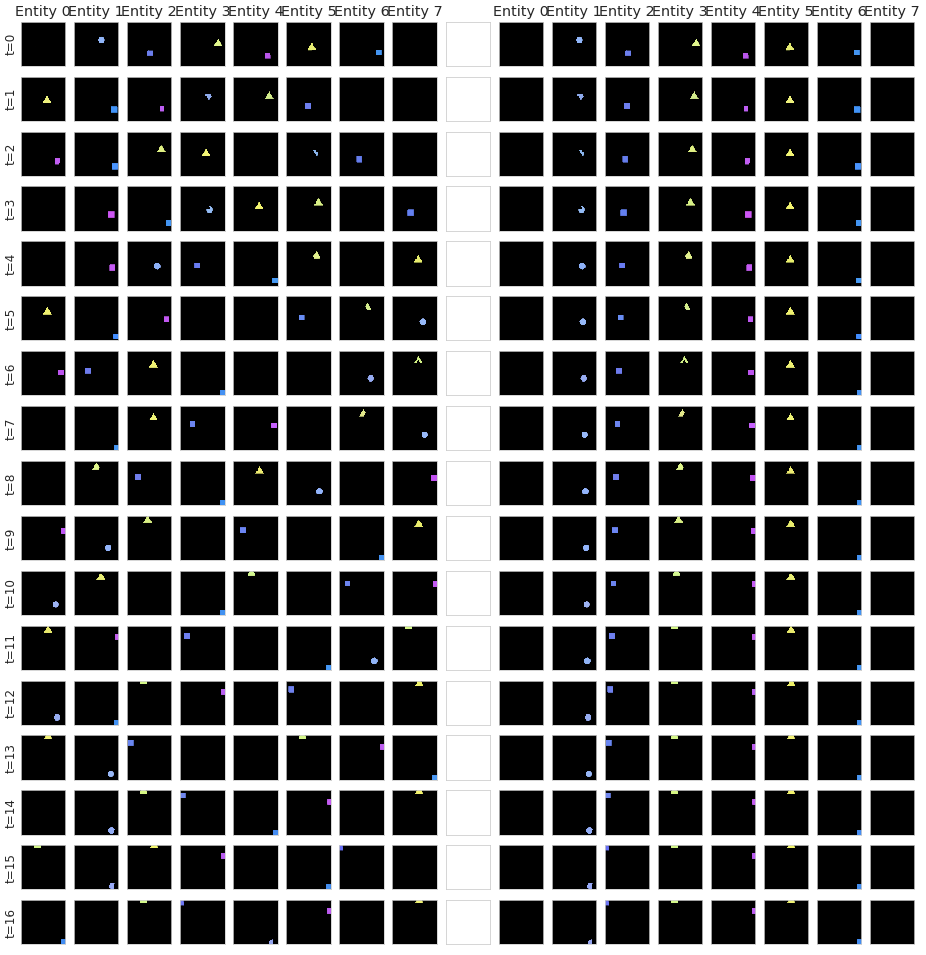}
    \caption{Visual results on SpriteWorld Task (a) (in Figure \ref{fig:sw:const_v}).  On the left of we show the inputs to the model and on the right the aligned entities. }
    \label{fig:SW_task_a_results}
\end{figure}

\begin{figure}
\begin{subfigure}[c]{0.24\textwidth}
    \centering
    \includegraphics[width=\textwidth]{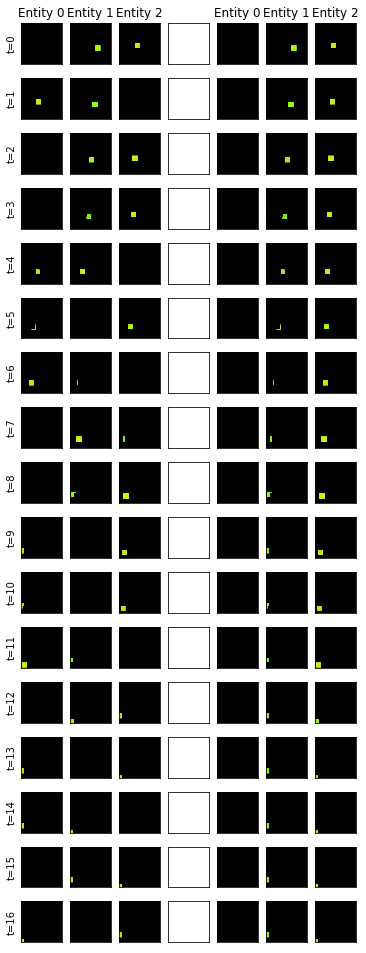}
    \caption{}
\end{subfigure}
\begin{subfigure}[c]{0.24\textwidth}
    \centering
    \includegraphics[width=\textwidth]{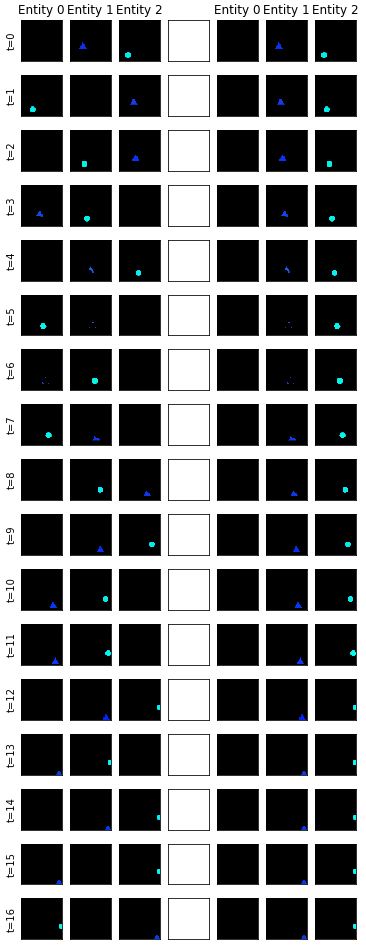}
    \caption{}
\end{subfigure}
\begin{subfigure}[c]{0.24\textwidth}
    \centering
    \includegraphics[width=\textwidth]{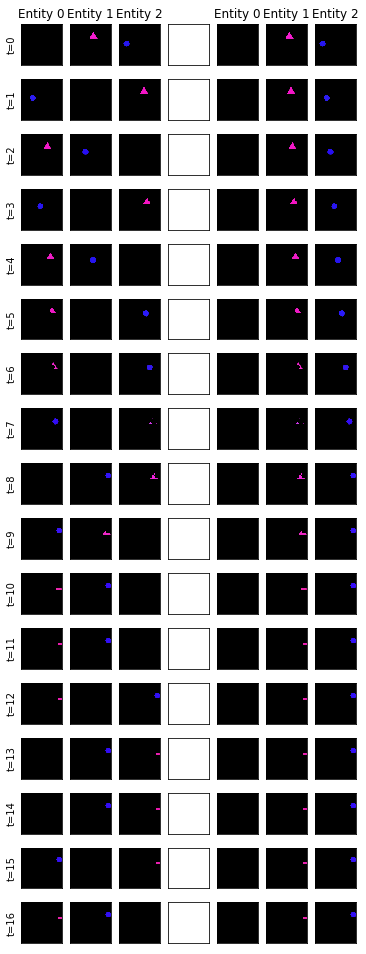}
    \caption{}
\end{subfigure}
\begin{subfigure}[c]{0.24\textwidth}
    \centering
    \includegraphics[width=\textwidth]{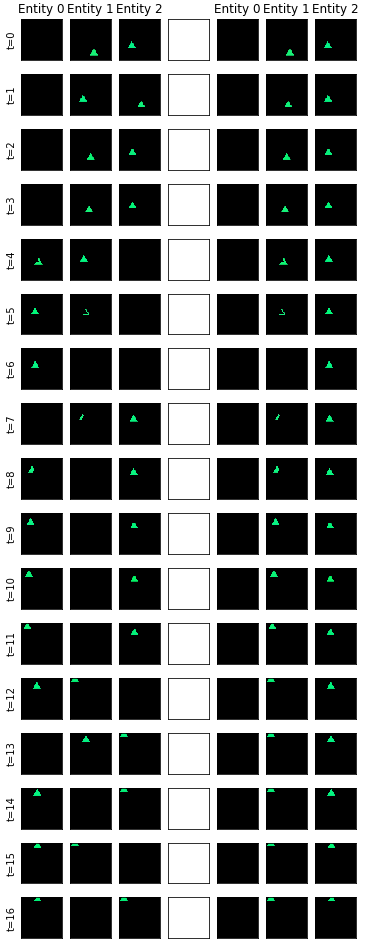}
    \caption{}
\end{subfigure}
\caption{Visual results on SpriteWorld Task (b) (Figure \ref{fig:sw:const_v_amb}). On the left of each sub-figure we show the inputs to the model and on the right the aligned entities. Sub figures (a) and (d) show results where both of the entities are the same colour and shape and they collide.}
\label{fig:SW_task_b_results}
\end{figure}

\begin{figure}
    \centering
    \includegraphics[width=\linewidth]{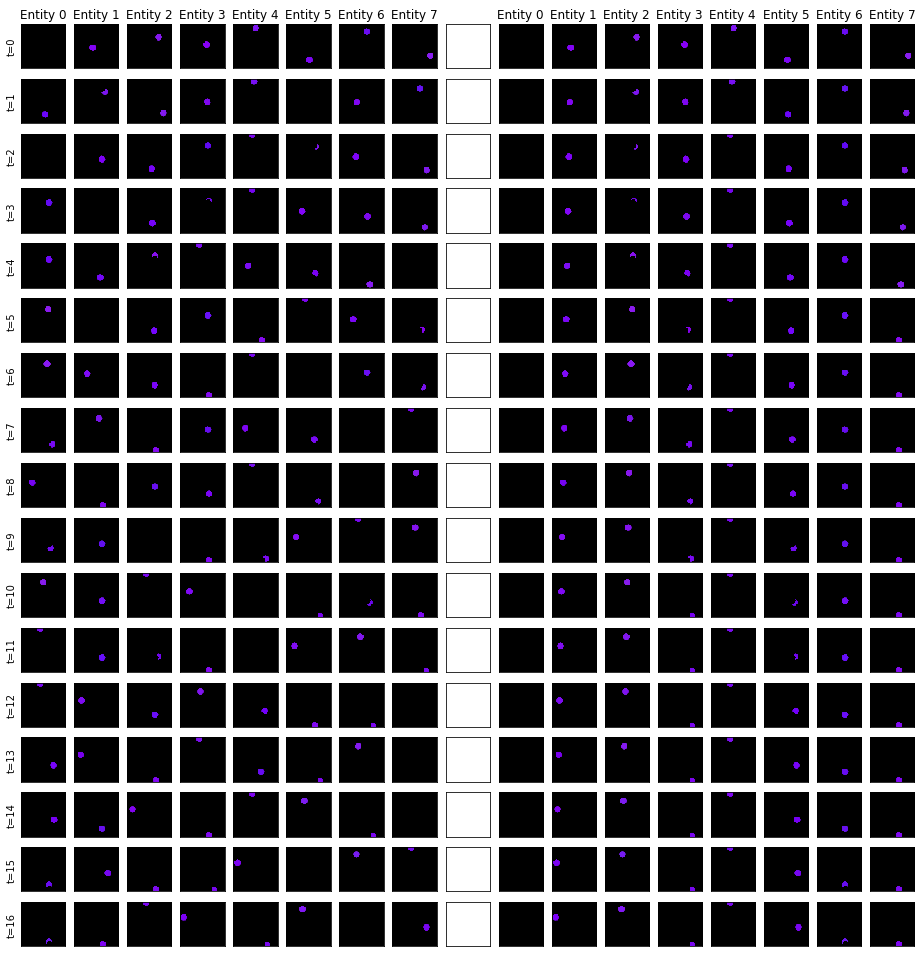}
    \caption{Visual results on SpriteWorld Task (c) (Figure \ref{fig:sw:const_v_same}.  On the left of we show the inputs to the model and on the right the aligned entities.}
    \label{fig:SW_task_c_results}
\end{figure}

\begin{figure}
    \centering
    \includegraphics[width=\textwidth]{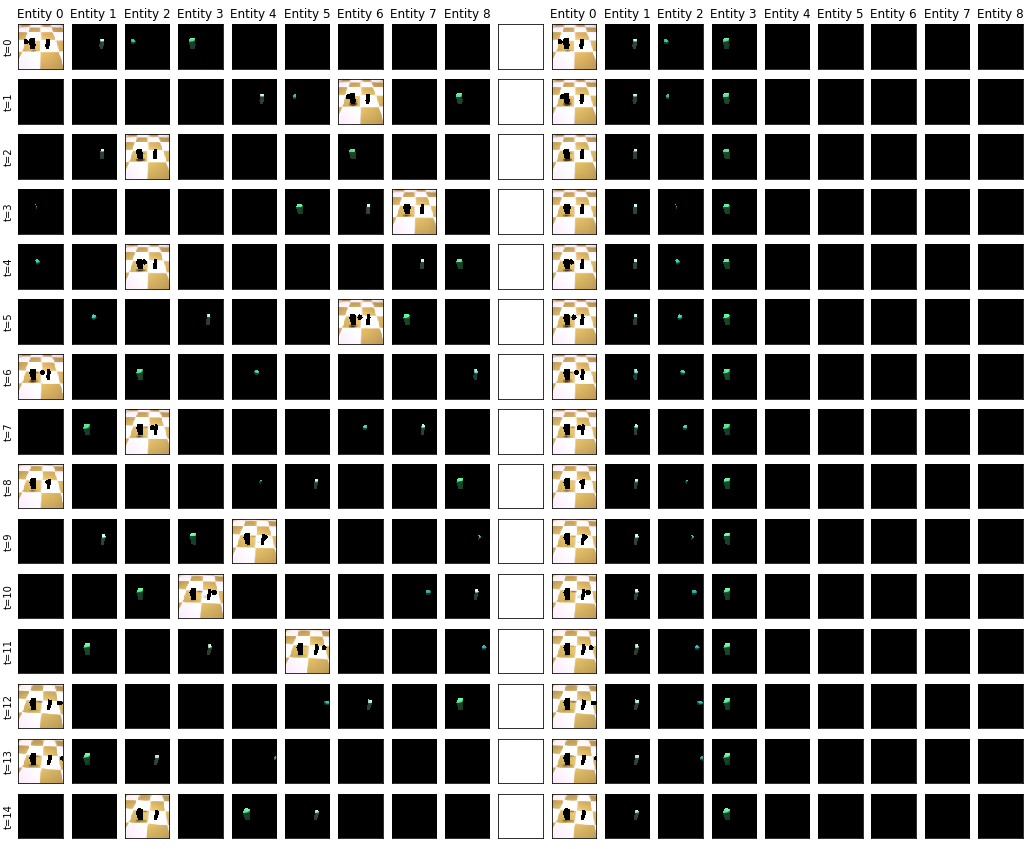}
    \caption{Visual results on the Physical Concepts: Continuity task. On the left are the inputs to the AlignNet on the right are the aligned outputs. Notice at $t=2$ that the ball becomes fully occluded, but at $t=3$ when the ball (Entity $2$) reappears it is assigned to the correct slot.}
    \label{fig:blai_results}
\end{figure}

\begin{figure}
    \centering
    \begin{subfigure}[c]{\textwidth}
        \includegraphics[width=\textwidth]{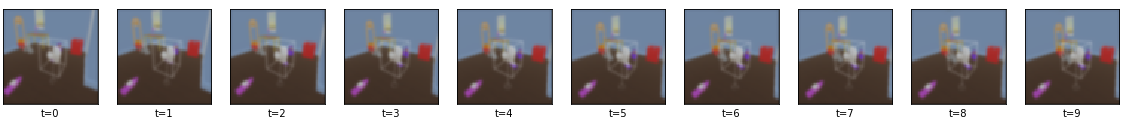}
    \caption{The inputs video sequence before applying MONet.}
    \end{subfigure}
    \begin{subfigure}[c]{\textwidth}
        \includegraphics[width=\textwidth]{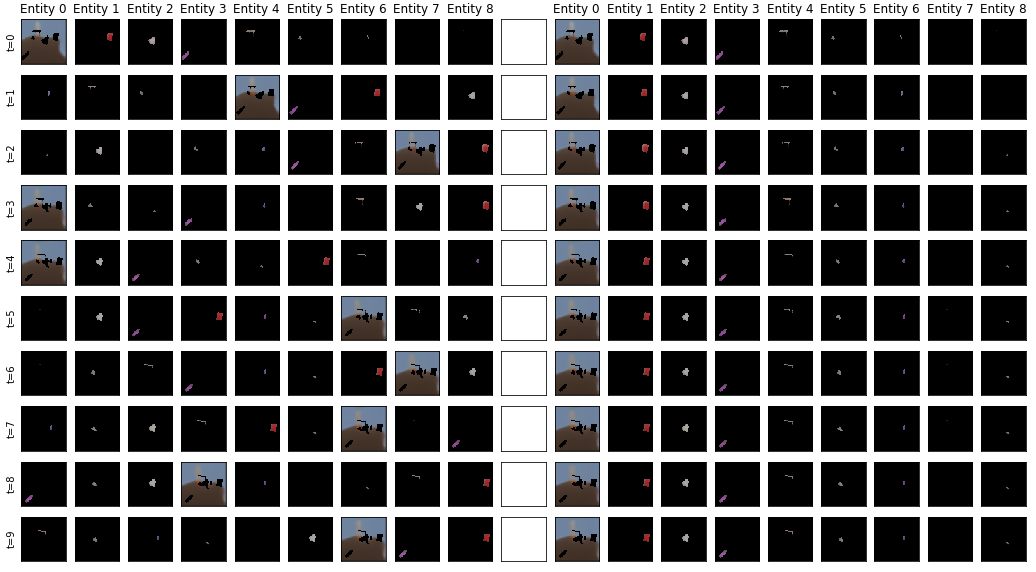}
    \caption{On the left the input seen by the AlignNet, on the right the output of the AlignNet.}
    \end{subfigure}
    \caption{Visual results on the Unity Room task.}
    \label{fig:room_results}
\end{figure}

\begin{figure}
    \centering
    \begin{subfigure}[c]{\textwidth}
        \includegraphics[width=\textwidth]{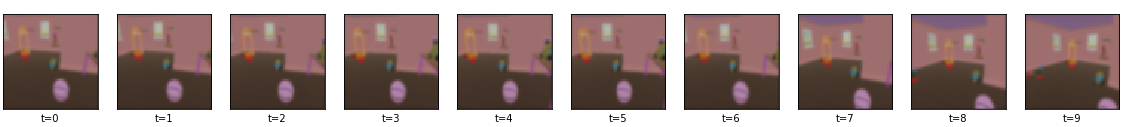}
    \caption{The inputs video sequence before applying MONet.}
    \end{subfigure}
    \begin{subfigure}[c]{\textwidth}
        \includegraphics[width=\textwidth]{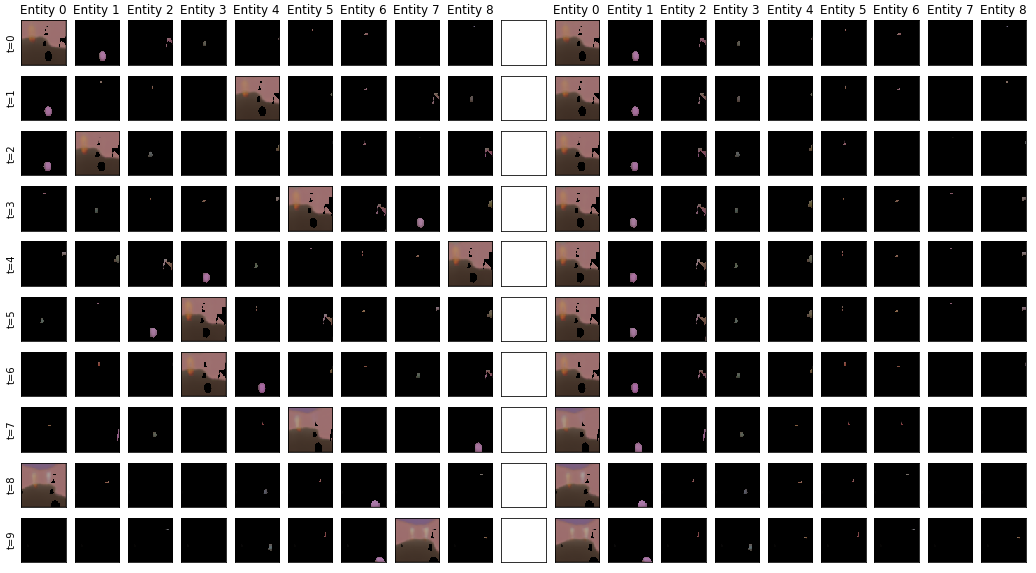}
    \caption{On the left the input seen by the AlignNet, on the right the output of the AlignNet. On the right, `Entity 2' is initially assigned a table, but once the table disappears at $t=7$ it is replaced by a similarly coloured object.}
    \end{subfigure}
    \caption{Visual results on the Unity Room task, a failure case.}
    \label{fig:room_results_fail}
\end{figure}

\end{document}